%% file: paper.tex
\documentclass{article}
\usepackage{amsfonts}
\usepackage{amssymb}
\usepackage{amsmath}
\usepackage{microtype}
\usepackage{graphicx}
\usepackage{subfigure}
\usepackage{booktabs} % for professional tables
\usepackage{hyperref}
\usepackage{algorithm}
\usepackage{tikz}
\usetikzlibrary{calc}
\usetikzlibrary{patterns}
\usetikzlibrary{shapes.multipart, arrows, positioning,shapes.geometric}
\usetikzlibrary{decorations.pathreplacing}
\usetikzlibrary{calc,shapes,decorations,decorations.shapes,positioning,shadows,arrows,decorations.markings,backgrounds,fit}
\usepackage{url}

\usepackage[accepted]{icml2018}

\icmltitlerunning{Flipped-Adversarial AutoEncoders}

\def\codename{$\textsc{f-AAE}$}

\begin{document}

\twocolumn[
\icmltitle{Flipped-Adversarial AutoEncoders}

\begin{icmlauthorlist}
\icmlauthor{Jiyi Zhang}{soc}
\icmlauthor{Hung Dang}{soc}
\icmlauthor{Hwee Kuan Lee}{bii,ipal}
\icmlauthor{Ee-Chien Chang}{soc}
\end{icmlauthorlist}

\icmlaffiliation{soc}{School of Computing, National University of Singapore.}
\icmlaffiliation{bii}{Bioinformatic Institute, Agency for Science, Technology and Research(A*STAR), Singapore.}
\icmlaffiliation{ipal}{Image and Pervasive Access Lab (IPAL), Singapore.}

%\icmlcorrespondingauthor{Jiyi Zhang}{zhangjiyi@u.nus.edu}
\icmlcorrespondingauthor{Hung Dang}{hungdang@comp.nus.edu.sg}

\icmlkeywords{Machine Learning, ICML}

\vskip 0.3in
]

\printAffiliationsAndNotice{} % otherwise use the standard text.
\definecolor{desertsand}{rgb}{0.93, 0.79, 0.69}
\definecolor{ashgrey}{rgb}{0.7, 0.75, 0.71}
\definecolor{bisque}{rgb}{1, 0.95, 0.87}

\input{abstract}
\input{intro}

\input{background}

\input{approach}

\input{eval}

\input{related_work}

\input{conclusion}

\bibliography{paper}
\bibliographystyle{icml2018}

\end{document}

%% file: abstract.tex
\begin{abstract}
%Generative Adversarial Networks (GANs) framework provides an efficient estimation procedure to construct a generative model . Nevertheless, GANs lack a mechanism to derive an ``inverse'' of the generative model; i.e., an encoder that. While there exists proposals that tackle this ``inverse mapping'' problem in the literature, their generators produce either ``oversmoothened'' images, or those that vaguely resemble the original images. In this paper, 

%An AutoEncoder simultaneously trains a generative model $G$  that maps an arbitrary latent code distribution to a data distribution, and its inverse, an encoder $E$. Recently, a few approaches were proposed to leverage adversarial training criterion in constructing autoencoders, notably Adversarial AutoEncoder (AAE) and BiGAN.   In this paper, we propose \textit{flipped-Adversarial AutoEncoder} (\codename) that flips the error objective in AAE -- it minimizes re-encoding errors in the latent space and exploits adversarial criterion in the data space.  Experimental evaluations on various standard image datasets demonstrate that such subtle flipping enables \codename~to produce sharper images, and to capture rich semantic representation of data. 

We propose a flipped-Adversarial AutoEncoder (\codename) that simultaneously trains a generative model $G$ that maps an arbitrary latent code distribution to a data distribution and an encoder $E$ that embodies an ``inverse mapping'' that  encodes a data sample into a latent code vector. Unlike previous hybrid approaches that leverage adversarial training criterion in constructing autoencoders, \codename~minimizes re-encoding errors in the latent space and exploits adversarial criterion in the data space.  Experimental evaluations demonstrate that the proposed framework produces sharper reconstructed images while at the same time enabling inference that captures rich semantic representation of data.

\end{abstract}

%% file: intro.tex
\section	{Introduction}
\label{sec:intro}

Learning  \textit{generative} models that capture the rich semantic representation of data  has been long considered a key challenge in the field of machine learning~\cite{importance_semantic}. Generative models, or \textit{generators} for short, are tasked to learn structured probability distributions, specifying a hypothetical casual process by which data are to be generated from latent structure~\cite{topics_semantic}. In another words, a generator maps  an arbitrary latent code distribution to the data distribution, and has to be learned from a sparse training set. Early approaches to construct generative models were typically based on maximum likelihood estimation, requiring approximating many intractable probabilistic computations~\cite{generative_DBM}. 
More recently, Goodfellow et al. proposed Generative Adversarial Networks (GAN) as an alternative technique to learn  generative models~\cite{gan}, embodying an adversarial training criterion so as to sidesteps difficulties faced by earlier approach. More specifically, the generator is pitted against a \textit{discriminative} model (a.k.a. \textit{discriminator}) which attempts to discriminate samples  drawn from the training data, from data generated by generator. The GAN framework has been empirically shown to obtain impressive results on natural images~\cite{dcgan, denton2015deep}.

Beside the generator,   there are also strong interests in its inverse,  that is,  encoder that maps the data distribution to the latent code distribution.  Such a pair of generator and encoder  would enable a wider range of applications, for instances,  feature representation,  noise removal via reconstruction, etc.   An autoencoder, in contrast to GAN, learns both generator and encoder simultaneously and there are extensive work on autoencoder construction \cite{vae}.
In view of the effectiveness of GAN, many have considered leveraging adversarial training criterion to learn the encoder, notably Adversarial AutoEncoder (AAE)~\cite{aae} and  BiGAN~\cite{bigan}. While these frameworks provide mechanisms to simultaneously train a pair of generator (or a decoder in AAE's terminology) and encoder altogether, thus addressing the problem at hand to some extent, each has its own limitations. In particular, images generated by AAE tend to be overly ``smoothened'', while those generated by BiGAN, although appear sharper, often vaguely resemble the original images.

\input{figures/figure_FAAE}

In this paper, we propose \textit{flipped-Adversarial AutoEncoder} (\codename). Our framework is inspired by AAE~\cite{aae}. Nevertheless, we introduce subtle yet significant modifications: flipping the positions of the encoder $E_\phi$ and generator $G_\theta$ in the training pipeline, and operating the discriminator $D_w$ in data space. Figure \ref{fig:FAAEarchitecture} depicts the architecture of our proposed framework. \codename~feeds a latent code vector $\textbf{z}$ drawn from a prior distribution $p_z(\textbf{z})$ into the generator $G_\theta$ and sample $\textbf{x}$ drawn from the data distribution $p_x(\textbf{x})$ into the discriminator $D_w$. The output $\hat{\textbf{x}}$ of $G_\theta$ is then pipelined to $E_\phi$, which re-encodes $\hat{\textbf{x}}$ back into a latent code vector $\hat{\textbf{z}}$. The generated image $\hat{\textbf{x}}$ is also fed into $D_w$, which attempts to discriminate $\hat{\textbf{x}}$ from $\textbf{x}$. 

While \codename~is similar to AAE in the use of dual objectives of  an \textit{error criterion} and an \textit{adversarial criterion}, the above-mentioned flipping leads to two key differences. Firstly, our framework measures and attempts to minimize the error (which we called {\em re-encoding error}) in the latent space, as opposed to AAE minimizing the {\em reconstruction error} between real and reconstructed images measured in the data space under some distance function (e.g. $\ell_2$ norm). The rational of focusing on re-encoding error is that,  latent space arguably carries semantic information. As such, optimising the error in latent space would minimise  perturbation of semantic representation of data. Secondly, the discriminator $D_w$ in our framework operates in the data space, as opposed to that of AAE operating in the latent space. This adversarial learning criterion enables \codename~to train the generator $G_\theta$ to produce samples as close to the data distribution $p_x$ as possible, overcoming the ``smoothening'' effect of AAE.

Our experimental evaluations demonstrate the advantages \codename~has over AAE and BiGAN. Figure \ref{fig:reconstruct-imagenet} shows the reconstructed images produced by  AAE, BiGAN and \codename, respectively, on the ImageNet dataset.  We observe that  \codename~tends to produce sharper images compared to AAE, supporting the rational of operating the  discriminator in the data space. We also find that the images generated by \codename~visually resemble the original  real images better than BiGAN's reconstructions, suggesting that \codename~attains generator and encoder that are closer to the inverse of each other. We suspect that this is  because  BiGAN, unlike AAE and \codename, does not explicitly minimise reconstructions error.  Instead, the framework  attempts to match two joint distributions in the  data and latent space, which indirectly gives  encoder and generator that are inverse of each other. 
However, in practice, it is arguably difficult to  meet the matching condition, and in such sub-optimal situation, it is not clear whether the reconstruction error would be small, even if the two joint distributions are  close.

In summary, our paper makes the following contributions:
\begin{enumerate}
\item  We propose a novel framework, namely flipped-Adversarial AutoEncoder (\codename), to simultaneously train a generator $G_\theta$, and an encoder $E_\phi$ considered as an ``inverse'' mapping of $G_\theta$. \codename~objective minimizes re-encoding errors in the latent space and exploits adversarial criterion in the data space. 

\item We conduct intensive experiment studies on standard datasets to empirically demonstrate that \codename~could produce high-quality samples, and at the same time enable inference that captures rich semantic representation of data. 
\end{enumerate}

%% file: figures/figure_FAAE.tex
\tikzset{latent/.style={circle,draw,minimum size=20pt}}

\tikzset{sample/.style={rectangle,draw,minimum size=25pt}}

\tikzset{triangle/.style={fill=desertsand, thick, regular polygon, regular polygon sides=3, minimum size=2pt}}
\tikzset{whitetriangle/.style={fill=white, thick, regular polygon, regular polygon sides=3, minimum size=2pt}}
\tikzset{white/.style={fill=white, regular polygon, regular polygon sides=4, minimum size=25pt}}

\tikzset{output/.style={fill = white, draw, thick, regular polygon, regular polygon sides=8, inner sep = -1pt}}

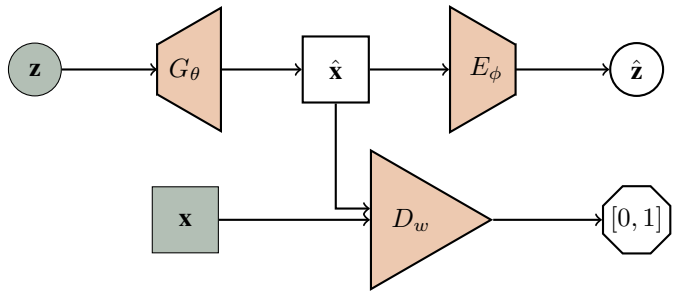
\begin{figure*}[t]
\centering      
%\begin{tikzpicture}[-,draw=black, node distance=\layersep,transform shape,rotate=90,scale=0.7]  
\begin{tikzpicture}[scale=0.75]  

\node[latent, fill = ashgrey](input_z){\textbf{z}};
\node[draw, triangle, shape border rotate = 90, right of = input_z, node distance = 2cm](G){$G_\theta$};
\draw[fill = white, draw = none] ($(G.west) + (0, -0.56)$) rectangle ($(G.west) + (0.8, 0.56)$);
\draw[thick] ($(G.west) + (0.8, -0.465)$) -- ($(G.west) + (.8, 0.465)$);

\node[sample, thick, right of = G, node distance = 2cm](gen_image){$\hat{\textbf{x}}$};
\node[draw, triangle, shape border rotate = -90, right of = gen_image, node distance = 2cm](E){$E_\phi$};
\draw[fill = white, draw = none] ($(E.east) + (-.80, -0.56)$) rectangle ($(E.east) + (0, 0.56)$);
\draw[thick] ($(E.east) + (-.80, -0.465)$) -- ($(E.east) + (-.80, 0.465)$);
\node[latent,thick, right of = E, node distance = 2cm](output_z){$\hat{\textbf{z}}$};
\node[sample, below of = G, node distance = 2cm, fill = ashgrey](input_image){\textbf{x}};
\node[draw, triangle, shape border rotate = -90, right of = input_image, node distance = 3cm](D){$D_w$};
\node[output, right of = D, node distance = 3cm](output){$[0,1]$};

\draw[->, thick]  (input_z.east) -- ($(G.west) + (0.8, 0)$){};
\draw[->, thick]  (G.east) -- (gen_image.west){};
\draw[->, thick]  (gen_image.east) -- ($(E.west)$){};
\draw[->, thick]  ($(E.east) + (-.80,0)$) -- (output_z.west){};
\draw[->, thick]  (input_image.east) -- (D.west){};
\draw[->, thick]  (D.east) -- (output.west){};
\draw[->, thick] (gen_image.south) |- ($(D.west)+ (0,0.2)$){};

\end{tikzpicture}
\caption{The proposed \codename~framework. A latent code vector $\textbf{z} \sim p_z(\textbf{z})$ is fed into the generator $G_\theta$, while an image $\textbf{x} \sim p_x(\textbf{x})$ is fed into the discriminator $D_w$. The output $\hat{\textbf{x}}$ of $G_\theta$ is pipelined to $E_\phi$, which re-encodes $\hat{\textbf{x}}$ into a latent code vector $\hat{\textbf{z}}$. The generated image $\hat{\textbf{x}}$ is also fed into $D_w$, which attempts to discriminate $\hat{\textbf{x}}$ from $\textbf{x}$.}
\label{fig:FAAEarchitecture}
\end{figure*}

%% file: background.tex
\section{Background}
\label{sec:background}

\subsection{Notations}
Let us first introduce some common variables and operators.
Variables include latent code vectors $\textbf{z} \in \mathbb{R}^n$ sampled from a known 
distribution $p_z(\textbf{z})$, generated 
latent code vectors $\hat{\textbf{z}} \in \mathbb{R}^n$,
data points $\textbf{x} \in \mathbb{R}^d$ 
drawn from the  prior data distribution $p_{x}(\textbf{x})$ and 
generated data points $\hat{\textbf{x}} \in \mathbb{R}^d$. When it is clear from the context, we abuse the notation, denoting $p_z(\textbf{z})$ by $p_z$ and $p_x(\textbf{x})$ by $p_x$. 

The operators include the {\it generator} (a.k.a. {\it decoder}) $G_\theta : \mathbb{R}^n \mapsto \mathbb{R}^d$, the {\it encoder} $E_\phi : \mathbb{R}^d \mapsto \mathbb{R}^n$ and the {\it discriminator} $D_w$. The generator $G_\theta$ takes as input either $\textbf{z}$ or $\hat{\textbf{z}}$, and outputs a generated data point $\hat{\textbf{x}}$. Thus, it plays a similar role to that of a generative network in the original GANs~\cite{gan}. The encoder $E_\phi$ takes as input either a real data point $\textbf{x}$ or a generated data point $\hat{\textbf{x}}$, and outputs a encoded latent vector $\hat{\textbf{z}}$. Finally, the discriminator $D_w$ takes in a pair of inputs, and is tasked to discriminate one from another. There are different functional forms of discriminator in different frameworks, which we shall elaborate in respective sections. The operators in this paper are all neural networks parameterized by $\theta, \phi, w$ for the generator, encoder and discriminator, respectively.

\subsection{GAN}
The GAN framework defines a min-max adversarial game which pits a generative model $G_\theta$ against a discriminative model $D_w$, as depicted in   Figure~\ref{fig:GAN_architecture}. The generator $G_\theta$ maps a latent code vector $\textbf{z} \sim p_z(\textbf{z})$, generating a sample $\hat{\textbf{x}}$. At the same time, the discriminator $D_w$  evaluates the probability that a given sample $\textbf{x}$ is drawn from the true data distribution $p_{x}(\textbf{x})$, instead of the generated by $G_\theta$. The optimal solution for this adversarial game is a pair of $G_\theta$ and $D_w$ that optimizes the  following function~\cite{gan}:

\begin{equation}
\begin{split}
\min_\theta \max_w  & \ {\cal A}(G_\theta, D_w)  \ \ \ \  \ \mbox{\em where} \\
{\cal A}(G_\theta, D_w) = \ & {\mathbb E}_{\textbf{x} \sim p_{x}}    [\log D_w(x)]  \\
 & + {\mathbb E}_{\textbf{z} \sim p_z} [\log (1 - D_w(G_\theta(z)))]
\end{split}
 \end{equation}
\input{figures/figure_GAN}

As mentioned in the previous section, GAN does not directly provide the encoder, that is, the inverse of the generator $G_\theta$.  Although one may apply generic method to  invert the generator $G_\theta$  produced by GAN, however,   it is not clear how to accurately and efficiently compute such inverse.  Moreover,  even if the inverse can be accurately derived from $G$,  the inverse mapping might not be continuous and smooth, and thus does not give a meaningful encoding that  captures  data semantic. 

\subsection{Adversarial AutoEncoders}
\input{figures/figure_AAE}
Adversarial AutoEncoder~\cite{aae} (AAE) framework introduces an additional constraint into the classical varational autoencoder (VAE)~\cite{vae} so as to enforce the ``hidden'' latent code  of the auto-encoder to observe statistical properties of some given prior distribution, thus guarantees that the hidden latent code  generated from any part of the data space would result in meaningful samples. More specifically, while VAE make leverages KL divergence penalty to impose a prior distribution on the hidden latent code  of the autoencoder, AAE employs an adversarial training to match the aggregated posterior with the prior distribution. The adversarial training procedure, similar in spirit to GAN, introduces a discriminative network $D_w: \mathbb{R}^n \mapsto \mathbb{R}$ into the training pipeline of the autoencoder. Note that unlike in the case of GAN, the discriminator for AAE maps the latent vector $\textbf{z}$ or $\hat{\textbf{z}}$ into the discriminator score.
$D_w$ is trained to distinguish a latent code  generated by the autoencoder from a real sample drawn from the prior distribution. The encoder is then trained to maximally confused $D_w$, while at the same time, optimized to minimize the reconstruction error of the autoencoder. 

From a prior distribution $p_{\bf z}$, the data distribution $p_{\bf x}$ and some distance function $d(\cdot, \cdot)$,
  AAE  jointly trains all three networks $G_\theta, E_\phi$ and 
$D_w$ by optimizing the following objective function (Figure \ref{fig:AAEarchitecture}):
\begin{equation}
\begin{split}
\min_{\theta,\phi } \max_w & \ {\cal B}  (G_\theta , E_\phi , D_w)  \ \ \ \  \ \mbox{\em where} \\
 {\cal B}(G_{\theta }, E_{\phi }, D_w)  =& \ {\mathbb E}_{\textbf{z} \sim p_{z}}  
                                                     [\log D_w(\textbf{z})]      \\ 
 & + \ {\mathbb E}_{\textbf{x} \sim p_{x}} [\log (1- D_w(E_\phi (\textbf{x}))]   \\ 
  & + \ {\mathbb E}_{\textbf{x} \sim p_{x}} [  d( \textbf{x}, G_\theta ( E_\phi (\textbf{x}) ) ) \ ]   
\end{split}
\end{equation}

The first two terms quantify the performance of the discriminator with respect to the encoder (that is, how well the prior distribution matches 
the aggregated posterior), and the last term is the {\em reconstruction error} incurred by the composition of encoder follows by generator.     A typical choice of the distance function is $\ell_2$ or $\ell_1$ norm. 

Note that in a solution with non-zero reconstruction error, the pair $G_\theta$ and $E_\phi$  are not the exact inverse of each other.   Indeed empirically, when applying to images, the reconstructed error are small  but  nonetheless non-zero.  Since the images within a bounded distance from the data are mostly  smoothened images, we would expect overly smoothened reconstructed images,  as depicted in  Figure \ref{fig:reconstruct-CIFAR-10} and \ref{fig:reconstruct-face}.

\subsection{BiGan}

\input{figures/figure_BiGAN}
BiGAN takes an interesting approach in learning the encoder.   The discriminator attempts to discriminate  joint  distributions in the data and latent space, that is, discriminating  $(G_\theta ({\bf z}), {\bf z})$  verse  $( {\bf x}, E_\phi ({\bf x}) )$.
The adversarial game optimises the following function:
\begin{equation}
\begin{split}
 \min_{\theta,\phi} \max_w &  \ {\cal C}  (G_\theta, E_\phi, D_w)  \ \ \ \  \ \mbox{\em where} \\
  {\cal C}(G_\theta, E_\phi, D_w ) =& \ {\mathbb E}_{\textbf{x} \sim p_{ data}} [\log   D_w(   E_\phi (x), x   )]  \\                         & +    {\mathbb E}_{\textbf{z} \sim p_z}  [ \log  (1-D_w (z, G_\theta (z) ))  ]    
\end{split}
\end{equation}

Although there is no explicit reconstruction lost term in the objective function,  it can be shown that with the  optimal encoder $E_{\phi^*}$  and generator $G_{\theta^*}$, the two joint distributions match,  which in turn implies $E_{\phi^*}$  is the  inverse of the  $G_{\theta^*}$ almost everywhere~\cite{bigan}.  However, in practice it is challenging to achieve optimality, and unfortunately,  it is not clear  whether the reconstruction error would be small at  near-optimal.  Empirically, for images,   $G_\theta ( E_\phi( x))$ is visually far from $x$, as depicted in Figure~\ref{fig:reconstruct-CIFAR-10} and \ref{fig:reconstruct-face}.

%% file: figures/figure_GAN.tex
\begin{figure}[t]
\centering    
\resizebox{.4\textwidth}{!}{  
\begin{tikzpicture}[scale=0.5]  
\node[latent, fill = ashgrey](input_z){\textbf{z}};
\node[draw, triangle, shape border rotate = 90, right of = input_z, node distance = 2cm, inner sep = 2pt](G){$G_\theta$};
\draw[fill = white, draw = none] ($(G.west) + (0, -0.6)$) rectangle ($(G.west) + (1, 0.56)$);
\draw[thick] ($(G.west) + (1, -0.57)$) -- ($(G.west) + (1, 0.57)$);
\node[sample, thick,right of = G, node distance = 2cm](gen_image){$\hat{\textbf{x}}$};
\node[sample, below of = G, node distance = 2.2cm, fill = ashgrey](input_image){\textbf{x}};
\node[draw, triangle, shape border rotate = -90, right of = input_image, node distance = 3cm, inner sep = 2pt](D){$D_w$};
\node[output, right of = D, node distance = 2cm](output){$[0,1]$};

\draw[->, thick]  (input_z.east) -- ($(G.west) + (1, 0)$){};
\draw[->, thick]  ($(G.east) + (0, 0)$) -- (gen_image.west){};
\draw[->, thick]  (input_image.east) -- (D.west){};
\draw[->, thick]  (D.east) -- (output.west){};
\draw[->, thick] (gen_image.south) |- ($(D.west)+ (0,0.2)$){};

\end{tikzpicture}
}
\caption{GAN Framework.  
 Unlike \codename, GAN only consists of $G_\theta$ and $D_w$. It lacks an encoder $E_\phi$ that maps a generated sample $\hat{\textbf{x}}$ into an encoded latent code vector $\hat{\textbf{z}}$, which can essentially serves as an inverse of $G_\theta$.}
\label{fig:GAN_architecture}
\end{figure}
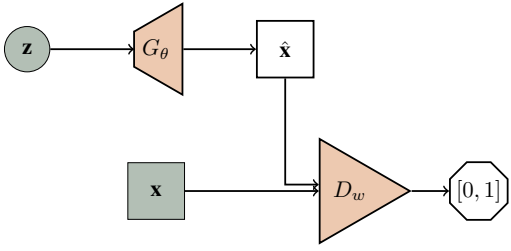

%% file: figures/figure_AAE.tex
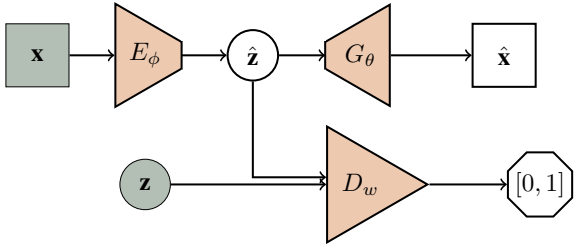
\begin{figure}[t]
\centering     
\resizebox{.45\textwidth}{!}{   
\begin{tikzpicture}[scale=0.5]  
\node[sample, fill = ashgrey](input_image){\textbf{x}};
\node[draw, triangle, shape border rotate = -90, right of = input_image, node distance = 1.5cm, inner sep = 2pt](E){$E_\phi$};
\draw[fill = white, draw = none] ($(E.east) + (-0.7, -0.5)$) rectangle ($(E.east) + (0.7, 0.5)$);
\draw[thick] ($(E.east) + (-0.7, -0.4)$) -- ($(E.east) + (-0.7, 0.4)$);

\node[latent, thick,right of = E, node distance = 1.5cm](gen_z){$\hat{\textbf{z}}$};
\node[draw, triangle, shape border rotate = 90, right of = gen_z, node distance = 1.5cm, inner sep = 2pt](G){$G_\theta$};
\draw[fill = white, draw = none] ($(G.west) + (0, -0.6)$) rectangle ($(G.west) + (0.7, 0.56)$);
\draw[thick] ($(G.west) + (.7, -0.4)$) -- ($(G.west) + (.7, 0.4)$);

\node[sample,thick, right of = G, node distance = 2cm](output_x){$\hat{\textbf{x}}$};
\node[latent, below of = E, node distance = 1.8cm, fill = ashgrey](input_z){\textbf{z}};
\node[draw, triangle, shape border rotate = -90, right of = input_z, node distance = 3cm, inner sep = 2pt](D){$D_w$};
\node[output, right of = D, node distance = 2.5cm](output){$[0,1]$};

\draw[->, thick]  (input_image.east) -- (E.west){};
\draw[->, thick]  ($(E.east) + (-0.7, 0)$) -- (gen_z.west){};
\draw[->, thick]  (gen_z.east) -- ($(G.west) + (.7, 0)$){};
\draw[->, thick]  (G.east) -- (output_x.west){};
\draw[->, thick]  (input_z.east) -- (D.west){};
\draw[->, thick]  (D.east) -- (output.west){};
\draw[->, thick] (gen_z.south) |- ($(D.west)+ (0,0.2)$){};

\end{tikzpicture}
}
\caption{AAE framework. In contrast to \codename, AAE feeds real data samples drawn from the data distribution $p_x(\textbf{x})$ into $E_\phi$, and exploits $D_w$ to impose a certain properties on encoded latent vector $\hat{\textbf{z}}$ that are generated by $E_\phi$.}
\vspace{-10pt}
\label{fig:AAEarchitecture}
\end{figure}

%% file: figures/figure_BiGAN.tex
\begin{figure}[t]
\centering    
\resizebox{.5\textwidth}{!}{   
\begin{tikzpicture}[scale=0.5]  
\node[latent, fill = ashgrey, inner sep = 2pt](input_z){\textbf{z}};
\node[latent, thick, below of = input_z, node distance = 3cm](output_z){$\hat{\textbf{z}}$};
\node[draw, triangle, shape border rotate = 90, right of = input_z, node distance = 1.5cm, inner sep = 2pt](G){$G_\theta$};
\draw[fill = white, draw = none] ($(G.west) + (-0, -0.56)$) rectangle ($(G.west) + (0.8, 0.56)$);
\draw[thick] ($(G.west) + (.8, -0.465)$) -- ($(G.west) + (.8, 0.465)$);
\node[sample, thick, right of = G, node distance = 1.5cm](gen_image){$\hat{\textbf{x}}$};
\node[draw, triangle, shape border rotate = 90, right of = output_z, node distance = 1.5cm, inner sep = 2pt](E){$E_\phi$};
\draw[fill = white, draw = none] ($(E.west) + (0.8, -0.56)$) rectangle ($(E.west) + (0, 0.56)$);
\draw[thick] ($(E.west) + (.8, -0.465)$) -- ($(E.west) + (.8, 0.465)$);

\node[sample, fill = ashgrey, right of = E, node distance = 1.5cm](input_image){\textbf{x}};
\node[draw, ellipse, fill = bisque, right of = gen_image, node distance = 1cm, yshift = -1.1cm, inner sep = 2pt] (a){$\hat{\textbf{x}}$, \textbf{z}};
\node[draw, ellipse, fill = bisque, right of = input_image, node distance = 1cm, yshift = 1.1cm, inner sep = 2pt] (b){\textbf{x}, $\hat{\textbf{z}}$};

\node[draw, triangle, shape border rotate = -90, right of = a, node distance = 1.35cm, yshift = -0.4cm, inner sep = 1pt](D){$D_w$};

\node[output, right of = D, node distance = 1.75cm](output){$[0,1]$};

\draw[->, thick]  (input_z.east) -- ($(G.west) + (.8, 0)$){};
\draw[->, thick]  ($(G.east)$) -- (gen_image.west){};
\draw[->, thick]  (D.east) -- (output.west){};
\draw[->, thick] (gen_image.east) -| ($(a.north)+(0, 0)$){};
\draw[->, thick] (input_z.south) |- ($(a.west)+(0, 0)$){};
\draw[->, thick]  (input_image.west) -- ($(E.east)$){};
\draw[->, thick]  ($(E.west) + (.8, -0)$) -- (output_z.east){};
\draw[->, thick] (input_image.east) -| ($(b.south)+(0, 0)$){};
\draw[->, thick] (output_z.north) |- ($(b.west)+(0, 0)$){};
\draw[->, thick]  (a.east) -- ($(D.west)+(0, 0.8)$){};
\draw[->, thick]  (b.east) -- ($(D.west)+(0, -0.8)$){};
\end{tikzpicture}
}
\vspace{-5pt}
\caption{BiGAN framework. As opposed to the \codename~discriminator, the discriminator of BiGAN receives joint pairs of latent code vector and data sample. In addition, BiGAN objective functions does not explicitly minimizes minimise reconstructions error. }
\label{fig:BiGAN}
\end{figure}
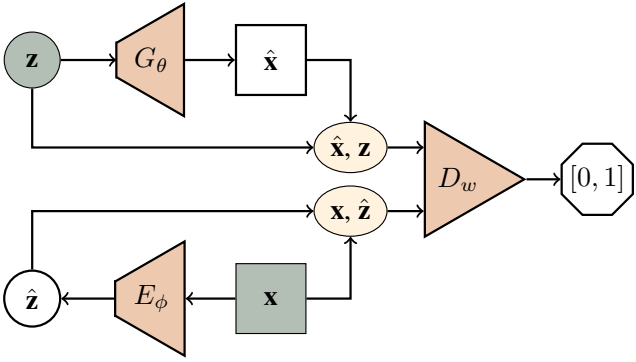

%% file: approach.tex
\section{Flipped Adversarial Auto Encoder (\codename)}
\label{sec:approach}

In this section, we present Flipped Adversarial Auto Encoder (\codename), 
contrasting our proposed framework against  BiGAN, 
AAE, and  Latent Vector Recovery. 

\subsection{Frame of \codename}

The inputs of \codename~training processes comprise latent vectors
$\textbf{z} \in \mathbb{R}^n$
from a known distribution $p_z(\textbf{z})$ and samples from the data
distribution $p_x(\textbf{x})$. Associated operators are the generator $G_\theta$, the 
encoder $E_\phi$ and the discriminators $D_w$ whose functions were explained in the previous section.

Our framework jointly trains all three networks $G_\theta, E_\phi$ and 
$D_w$ by optimizing the following objective function:
\begin{equation}
\begin{split}
\min_{\theta,\phi} \max_w  \ & {\cal F}  (G_\theta , E_\phi , D_w)  \ \ \ \  \ \mbox{\em where} \\
{\cal F}(G_{\theta }, E_{\phi }, D_w)  = &
              \ {\mathbb E}_{\textbf{x} \sim p_{x}}[\log D_w(\textbf{x})]      \\ 
& + \ {\mathbb E}_{\textbf{z} \sim p_{z}} [\log (1 - D_w(G_\phi (\textbf{z})))]   \\ 
& + \alpha  \ {\mathbb E}_{\textbf{z} \sim p_{z} }[  \| \textbf{z} - E_\theta (G_\phi (\textbf{z})) ) \| \ ]  
\end{split}
\end{equation}

The last term measures the difference after a latent vector $\textbf{z}$ is being ``re-encoded''. While our objective function adopts $\ell$-2 norm, other distance functions are also applicable. The pre-defined constant $\alpha$ serves as a normalizing factor  for the distance function.  There are many options for the latent vector distribution $p_z(\textbf{z})$.  In our experiments,  we sample the latent vector $\textbf{z} \sim p_z(\textbf{z})$ uniformly from a $n$ dimensional unit
sphere $S^n$; i.e., a multivariate normal distribution conditioned on unit norm.

The training iterates two phases, namely {\it re-encoding} and {\it regularization}:

\begin{itemize}
\setlength\itemsep{0em}

\item  Re-encoding phase:  $G_\theta$ and $E_\phi$ are updated so as to minimize the  re-encoding error between the input $\textbf{z}$ and the re-encoded latent vector $\hat{\textbf{z}} =E_\phi(G_\theta (\textbf{z}))$.

\item  Regularization phase: $D_w$ is first updated to discriminate the real sample $\textbf{x} \sim p_x(\textbf{x})$ from the generated sample $\hat{\textbf{x}} = G_\theta (\textbf{z})$.
Next, the generator $G_\theta$ is updated so as to maximally confuse the discriminator $D_w$. 
\end{itemize}

At the end of a successful training, we obtain a generator $G_\theta$ that is capable of 
generating meaningful samples resembling those that are drawn from the true data distribution.
In addition, we also obtain an encoder $E_\phi$ that approximates the  inverse of $G_\theta$ such that the re-encoding error is small.

\input{FAAE_construction}

\input{FAAE_vs_latent}

\input{FAAE_vs_AAE}

%% file: FAAE_construction.tex
\subsection{Architecture of $D_w$, $G_\theta$ and $E_\phi$}
%Our construction of the networks adopts the idea of Deep Convolutional Generative Adversarial Networks (DCGAN)~\cite{dcgan}.

The generator $G_\theta$ is a deconvolutional neural network which takes as input a vector from latent space. In our experiments, we represent such latent vector by a random array of dimension 256 sampled uniformly from a $n$-dimensional unit sphere. The output of $G_\theta$ has the same shape as the real sample training data; i.e., $32 \times 32 \times 3$ for CIFAR-10 and CelebA dataset, and $64 \times 64 \times 3$ for ImageNet dataset. 

Up-sampling layers are interleaved with convolution layers to reshape the input latent vector and scale it to the same dimension as training data. Batch Normalizaton~\cite{batch_norm} is used in between the layers to reduce covariate shift. A leaky ReLU layer is used after each Batch Normalizaton layer. The last layer uses a sigmoid function as the activation function to ensure the pixel value fall between 0 and 1 which is the same as our preprocessed and normalized dataset.

The encoder $E_\phi$ uses an almost identical architecture as the generator but in a reverse order. The up-sampling layers are replaced with max-pooling layers. As the end of the network, the feature map is flattened back to an array of the given latent dimension. The output shape of the encoder is the same as the input shape of the generator $G_\theta$ and the input shape of the encoder is the same as the output shape of the generator $G_\theta$. 

The discriminator $D_w$ is also a convolutional neural network. The input shape is the same as the output shape of the generator $G_\theta$. The output is a probability indicating how real the input is. Sigmoid function is used as the activation function for the last layer to ensure the output falls between $0$ and $1$.

%% file: FAAE_vs_latent.tex
\subsection{Relationship to Latent Vector Recovery}
There are other ways to implement a reverse mapping of Generative Adversarial Network (GAN) \cite{gan}. Lipton \& Tripathi at al. suggested the use of stochastic clipping which is a gradient-based technique to find the reverse function of an existing trained GAN generator~\cite{lipton2017precise}. \codename~, in contrast, does not attempt to reverse arbitrary function. Instead, our framework provides a method to constrain the training of the generator, guaranteeing the existence of an approximate inverse mapping of the generator at any time during the training. 

%% file: FAAE_vs_AAE.tex
\subsection{Relationship to Adversarial Autoencoder}
\codename~introduces a subtle yet significant modification over AAE's architecture, which is to flip the positions of  $E_\phi$ and $G_\theta$ in the training pipeline, and operating $D_w$ in data space (Figure \ref{fig:FAAEarchitecture}). This leads to two key differences. Firstly, \codename~measures and minimizes the re-encoding error in the latent space, as opposed to AAE minimizing reconstruction error in data space. The phenomenon of vector arithmetic on the latent vectors in GAN suggest that imposing measurement on the latent vectors can yield meaniful results~\cite{li_gan}. Secondly, operating $D_w$ in the data space allows \codename~to capture the data distribution $p_x(\textbf{x})$ directly via the adversarial criterion. These advantage enables \codename~the generator $G_\theta$ produce sharper images compared to that of AAE.

%% file: eval.tex
\section{Empirical Evaluation}
\label{sec:eval}
\input{figures/fig_re_cifar}

\input{figures/fig_re_face}

\input{figures/fig_re_imagenet}
We present three use cases of \codename~and benchmark it against AAE and BiGAN, which are the most related approaches to our work.  BiGAN presents an interesting alternative method of finding the inverse of image generation without explicitly calculating re-encoding or reconstruction loss. Comparison are performed on the three standard datasets which are CIFAR-10, ImageNet and CelebA. To enable reproducibility of our experiment results, we made a prototype of \codename~available online\footnote{\url{https://github.com/zhangjiyi/FAAE}}.

\subsection{Experiment Setup}
We used CIFAR-10, CelebA and ImageNet datasets for our training and testing. The CIFAR-10 contains $50K$ training samples and $10K$ testing samples. Each image has a shape of $32 \times 32 \times 3$. The CelebA dataset comprises $202599$ cropped and aligned face images. We resized the images to $32 \times 32 \times 3$, using five sixth of them for training and one sixth for testing. For ImageNet, we use the same dataset as Donahue et al.~\cite{bigan}, resizing the images to $64 \times 64 \times 3$. Our experiments normalize all the images so that each pixel takes a value in range $[0, 1]$. 

For the comparsion with AAE on CIFAR-10 and CelebA, we trained both \codename~and AAE for $50$ epoches. We set the same weightage of reconstruction/re-encoding loss and adversarial loss: $10^2$ for reconstruction/re-encoding loss and $0.1$ for adversarial loss. We used Adam optimizer and the same learning rate for \codename~and AAE: $3 \cdot 10^{-4}$ for generator and $10^{-3}$ for discriminator. We also use the same decay rate for these two models: $10^{-4}$ for both generator and discriminator.

Since we were not able to reproduce the BiGAN results, we restrict the benchmark against BiGAN to ImageNet dataset, taking the results from BiGAN paper directly and presenting them in Figure \ref{fig:reconstruct-imagenet}. In this set of experiments, we trained \codename~and AAE for $600$ epoches, using the same optimizer, learning rate and decay rate as above. For the weightage of reconstruction/re-encoding loss and adversarial loss, we used $30$ for reconstruction/re-encoding loss and $0.1$ for adversarial loss in the first $200$ epoches and $100$ for reconstruction/re-encoding loss and $0.1$ for adversarial loss for the rest epoches. The training of AAE and \codename~in our experiments always take the same parameters.

\input{figures/fig_gen_face}

\subsection{Image Reconstruction}

The first comparison is done on the image reconstruction tasks wherein images are encoded into latent space representations and then reconstructed back into the data space. Figure \ref{fig:reconstruct-CIFAR-10} and \ref{fig:reconstruct-face} show reconstructions of CIFAR-10 images using AAE and \codename.
AAE tends to produce blurry reconstructions whereas \codename~produce sharper images.

Note that for the CelebA data set, a majority of the training images are frontal photos of the faces. Since \codename~ is trained based on re-encoding errors measured in the latent semantic space, \codename~tends to reconstruct semantic representation, which is mostly frontal views for this dataset.  That is, given a side view face photo, \codename~has the tendency to ``turn" the face into a frontal view. Some examples of this phenomenon is highlighted with arrows in Figure \ref{fig:reconstruct-face}.

Figure \ref{fig:reconstruct-imagenet} shows reconstruction results in comparison with BiGAN and AAE. We find that BiGAN observes numerous reconstruction failures. For example, the eagles in the top first and second panels, the owl in the second panel and birds in the third panel cannot be reconstructed. While it can be proven that BiGAN converges to inverse mappings, it is unclear if conditions for obtaining convergence to global minimum could be met in actual training.

\input{figures/fig_morph}
\subsection{Image Generation}

Random ``meaningful" images can also be generated by feeding latent code vectors sampled in the latent space to the generator $G_\theta$. Figure \ref{fig:gen-face} depicts generated faces using the CelebA data set as the training set, with top panels are those produced by \codename~while bottom panel is generated using AAE. As expected, we find that images generated using AAE tend to be blurry.

\subsection{Image Morphing through vector arithmetic}
With good generator and encoder that are inverses of each other, image morphing can be performed by mapping
images into latent vectors, perturbing the latent vectors and mapping them back onto the data space.
Figure \ref{fig:morphing} demonstrate results of image morphing using {\codename} and AAE. The same experiments  are performed for both methods. In the image panels, four corner images are generated by manually choosing four original images, encoding them into latent space and reconstructing the images from the encoded latent vectors. Let the (un-normalized) encoded latent vectors be $\textbf{z}_1, \textbf{z}_2, \textbf{z}_3$ and $\textbf{z}_4$, our experiment generates their linear combinations as follow:
\begin{equation}
\textbf{l} = \alpha_1 \textbf{z}_1 + \alpha_2 \textbf{z}_2 + \alpha_3 \textbf{z}_3 + \alpha_4 \textbf{z}_4
\end{equation}
We then normalize $\textbf{z}' = \textbf{l}$ to generate a new latent vector that will be a sample from $p_z$.
Finally, we map $\textbf{z}'$ back into the data space, attaining $\textbf{x}'$. By gradually varying $\alpha_i$, this mechanism achieve an effect of morphing.

We plot the  morphed images produced by \codename and AAE in Figure \ref{fig:morphing}. We observe that
images generated by AAE (bottom panel) are blurry and visually less consistent. For example, the images in the center of the AAE panel seems appears more smoothened than those at the corners. \codename~generated images, on the other hand, are visually more consistent throughout the panel. We believe that this is by courtesy of operating \codename's discriminator in the data space.

%% file: figures/fig_re_cifar.tex
\begin{figure}[t]
\begin{picture}(0,130)
\put(0,0){\includegraphics[width=\linewidth]{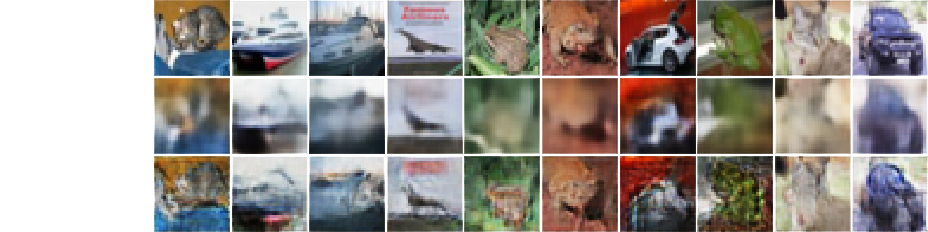}}
\put(0,70){\includegraphics[width=\linewidth]{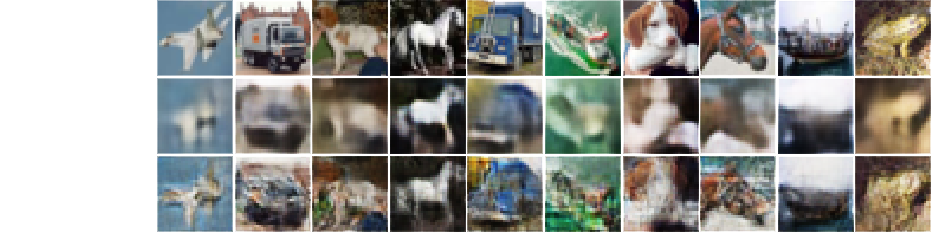}}
\put(8,7){\scriptsize {\codename}}
\put(08,27){\scriptsize AAE}
\put(08,47){\scriptsize Real}
%\put(08,40){\scriptsize originals}
\put(08,77){\scriptsize {\codename}}
\put(08,97){\scriptsize AAE}
\put(08,117){\scriptsize Real}
%\put(08,115){\scriptsize originals}
\end{picture}
\caption{Reconstruction comparison between AAE and {\codename} on CIFAR-10 dataset using an image resolution of 32x32.}
\vspace{-5pt}
\label{fig:reconstruct-CIFAR-10}
\end{figure}

%% file: figures/fig_re_face.tex
\begin{figure}[t]
\begin{picture}(0,130)
\put(0,0){\includegraphics[width=\linewidth]{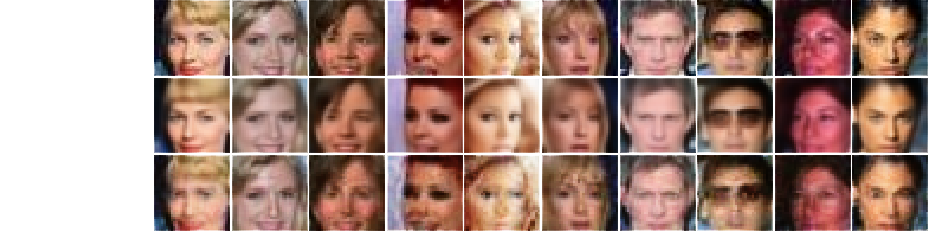}}
\put(0,70){\includegraphics[width=\linewidth]{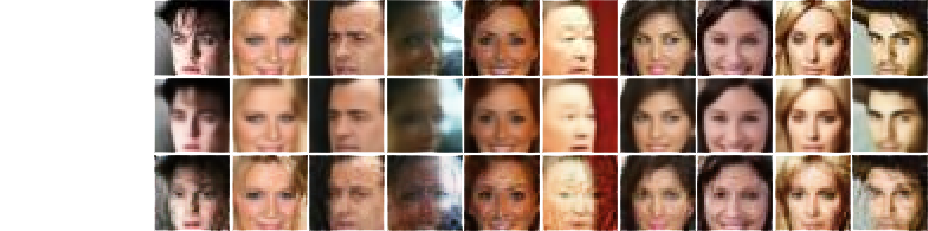}}
\put(08,7){\scriptsize {\codename}}
\put(08,27){\scriptsize AAE}
\put(08,47){\scriptsize Real}
\put(08,77){\scriptsize {\codename}}
\put(08,97){\scriptsize AAE}
\put(08,117){\scriptsize Real}
\put(138,0){\color{red} \framebox(18.5,19)}
\put(118,0){\color{red} \framebox(18.5,19)}
\put(98,70){\color{red} \framebox(18.5,19)}
\put(137,70){\color{red} \framebox(18.5,19)}
\put(145,-10){\vector(0,1){8}}
\put(145,60){\vector(0,1){8}}
\put(110,60){\vector(0,1){8}}
\put(125,-10){\vector(0,1){8}}
\end{picture}
\caption{Reconstruction comparison between AAE and {\codename} on CelebA dataset using an image resolution of 32x32.}
\vspace{-5pt}
\label{fig:reconstruct-face}
\end{figure}
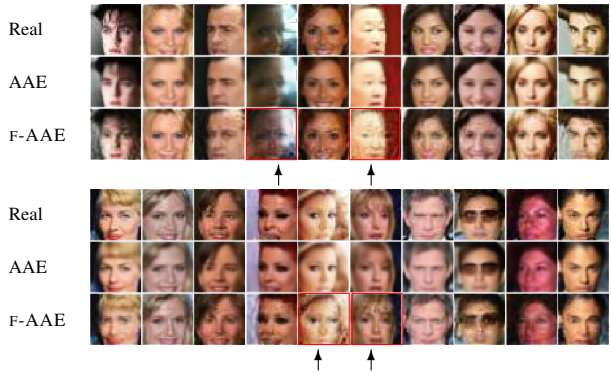

%% file: figures/fig_re_imagenet.tex
%\begin{figure*}[t]
%{\includegraphics[width=16cm]{figures/test_01-title}}
%{\includegraphics[width=16cm]{figures/test_02-title}}\\
%%{\includegraphics[width=13cm]{figures/test_03-title}}\\
%%{\includegraphics[width=13cm]{figures/test_04-title}}\\
%{\includegraphics[width=16cm]{figures/test_05-title}}\\
%{\includegraphics[width=16cm]{figures/test_06-title}}\\
%\caption{Comparison of reconstructed image $G_\theta(E_\phi ( \bf x)))$  between BiGAN, AAE and \codename for CIFAR10 dataset.
%BiGAN generally does not reconstruct well and AAE reconstructions are blurr as explained in the text.
%\codename reconstructs with some noise and artefacts.}
%\label{fig:reconstruct-imagenet}
%\end{figure*}

\begin{figure*}[t!]
\begin{picture}(0,600)
\put(0,0){\includegraphics[width=0.9\linewidth]{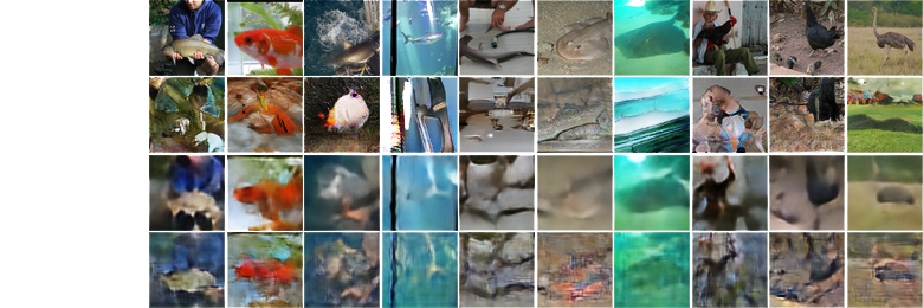}}
\put(0,150){\includegraphics[width=0.9\linewidth]{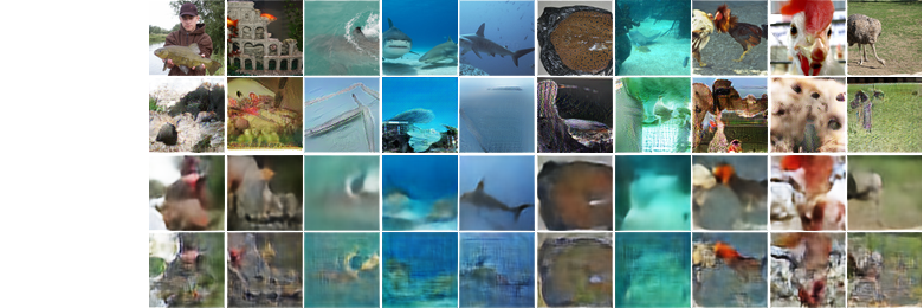}}
%\put(0,170){\includegraphics[width=\linewidth]{figures/test_03-title}}
%\put(0,255){\includegraphics[width=\linewidth]{figures/test_04-title}}
\put(0,300){\includegraphics[width=0.9\linewidth]{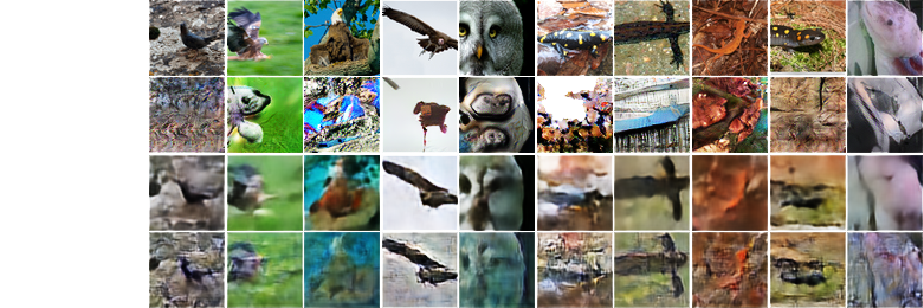}}
\put(0,450){\includegraphics[width=0.9\linewidth]{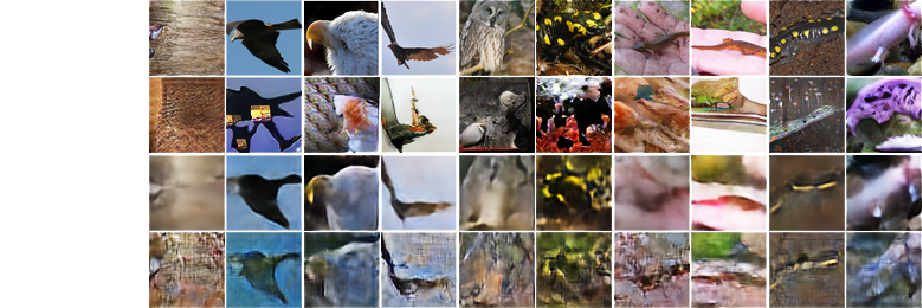}}
%\put(28,492){ ${\bf x}$}
%\put(28,472){ BiGAN}
%\put(28,452){ AAE}
%\put(28,432){ {\codename}}
%\put(28,407){ $\bf{x} $}
%\put(28,387){ BiGAN}
%\put(28,367){ AAE}
%\put(28,347){ {\codename}}
\put(28,575){ Real}
\put(28,538){ BiGAN}
\put(28,500){ AAE}
\put(28,465){ {\codename}}
\put(28,425){ Real}
\put(28,388){ BiGAN}
\put(28,350){ AAE}
\put(28,315){ {\codename}}
\put(28,275){ Real}
\put(28,238){ BiGAN}
\put(28,200){ AAE}
\put(28,165){ {\codename}}
\put(28,128){ Real}
\put(28,87){ BiGAN}
\put(28,52){ AAE}
\put(28,17){ \codename}
\end{picture}
\caption{Reconstruction comparison between BiGAN, AAE and \codename~on ImageNet dataset. BiGAN generally does not reconstruct well and AAE reconstructions are blurry. Images reconstructed using \codename~contain some noise and artifacts.}
\label{fig:reconstruct-imagenet}
\end{figure*}
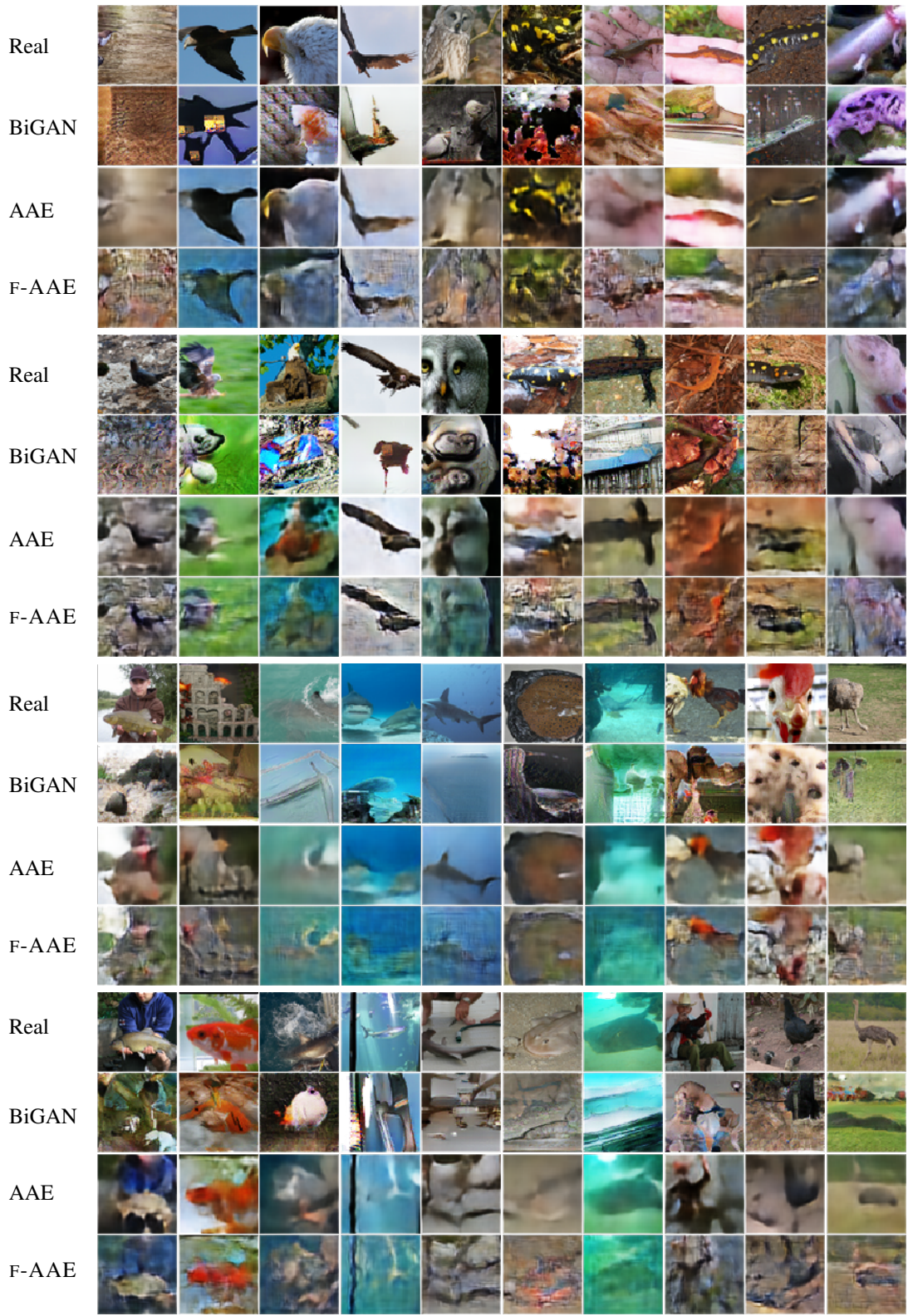

%% file: figures/fig_gen_face.tex
\begin{figure}[t]
\begin{picture}(0,250)
\put(0,0){\includegraphics[width=\linewidth]{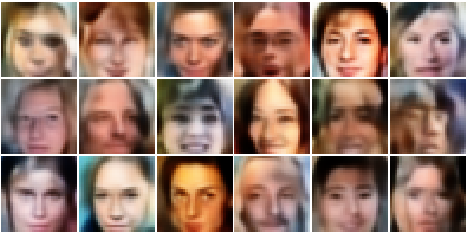}}
\put(0,130){\includegraphics[width=\linewidth]{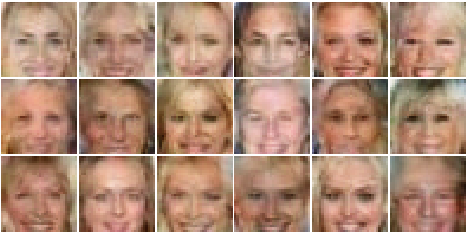}}
\end{picture}
\caption{Randomly generated faces by {\codename} (top panel) and AAE (bottom panel) trained for image resolution of 32x32.}
\vspace{-0pt}
\label{fig:gen-face}
\end{figure}

%% file: figures/fig_morph.tex
\begin{figure}[t!]
\begin{picture}(0,400)
\put(20,200){\includegraphics[width=0.85\linewidth]{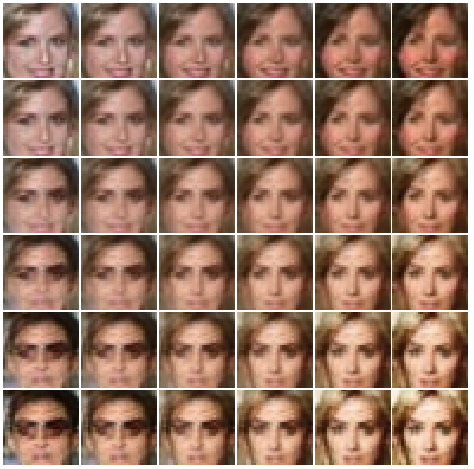}}
%\put(0,300){\Huge \color{white}REPLACE THIS W FAAE}
%\put(0,75){\Huge \color{white}with {\codename} results}
\put(20,0){\includegraphics[width=0.85\linewidth]{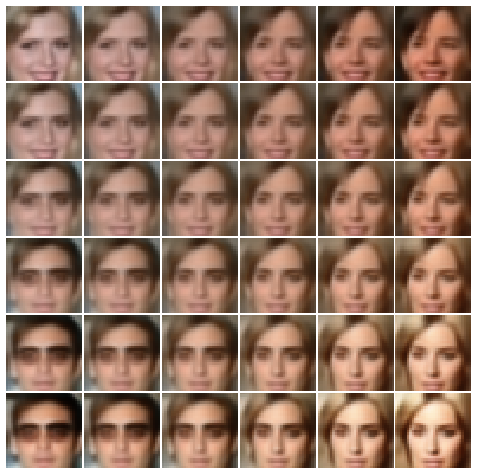}}
%\put(0,50){\Huge \color{white}MAKE TILE:5x5/6x6}
\end{picture}
\caption{{\codename} (top) and AAE (bottom) results for image morphing. Images at the four corners of each panel are reconstructions of real images. The rest were generated from linear combinations of the latent space representations of the real images.}
\vspace{-5pt}
\label{fig:morphing}
\end{figure}

%% file: related_work.tex
\section{Related Work}
\label{sec:related_work}
The GAN framework has inspired a vast body of research in the literature, focusing on either augmenting GANs with additional functionality or improving its performance. Huang et al. explored a generative model which is used to invert hierarchical representations of a discriminative network~\cite{stacked_gan}. Mirza et al. presented conditional GAN (cGAN), augmenting GANs such that the data generation is conditioned by both intrinsic (i.e., latent code vector) and extrinsic (i.e., known auxiliary information) factors~\cite{cgan}. 
Perarnau et al.  discussed techniques for learning inverse mappings of generators pretrained under cGAN ~\cite{icgan}. Our framework, on the other hand, focuses solely on learning the generator and its inverse mapping simultaneously, without constraining the data generation with any extrinsic factors or attempting to invert discriminative networks. 

Other related proposals that focus on training both the generator and the encoder under the GAN framework include BCGAN~\cite{bcgan}, BiGAN~\cite{bigan}, and adversarially learned inference (ALI) model~\cite{ali}. While \codename~shares with these proposals the use of adversarial training criterion, our framework operates the discriminator in a different fashion. In particular, while the discriminators in the above mentioned proposals receive joint pairs of latent code vector and data sample ($\textbf{x}, \textbf{z}$), the discriminator in \codename~operates solely in the data space. Besides, BiGAN and ALI do not explicitly minimizing reconstruction errors, as opposed to \codename~does. 

Perhaps most closly related to \codename~is the Adversarial AutoEncoder (AAE)~\cite{aae}. AAE is an extension of the Variational Autoencoders (VAE)~\cite{vae}, and thus suffering from VAE's limitation wherein samples generated by the decoder are blurry and overly smoothened.  \codename~overcomes this limitation by pitting $G$ against a discriminator $D$ operating in the data space, thus enabling the generator $G$ to produce samples as close to the data distribution $p_x$  as possible (i.e., attaining higher visual quality). Another point contrasting AAE and \codename~is a mechanism that encourages the encoder distribution to match with the prior distribution of the latent code vector. \codename~attains this by imposing mean square error criterion on latent code vectors, instead of AAE's use of the discriminator operating in the latent space.

%% file: conclusion.tex
\section{Conclusion and Future Work}
\label{sec:conclusion}
We have presented the flipped-Adversarial AutoEncoder that simultaneously trains a generative model $G$ that maps arbitrary latent code distribution to the data distribution, and an encoder $E$ that can be considered as an ``inverse'' mapping of $G$. The key technical novelty of \codename~lies in its optimization strategy which minimizes re-encoding errors in the latent space and exploits adversarial criterion in the data space. We have empirically demonstrated on various standard datasets that \codename~could produce high-quality samples, and at the same time enable inference that captures rich semantic representation of data. 

The proposed \codename~framework potentially enables a range of interesting applications. We hypothesize that a good pair of generator and decoder that allow bidirectional mapping between a complex data distribution and a simpler latent distribution can be used to learn underlying image manifold via differential
latent vector arithmetic. In particular, given an image $\textbf{x}$, one can employs \codename's encoder to encode $\textbf{x}$ into $\textbf{z}$. A small perturbation can then be applied to the latent vector $\textbf{z}$, obtaining $\textbf{z}' = \textbf{z}+\Delta \textbf{z}$ which can be mapped (by the generator $G$) back onto a morphed image $\textbf{x}'$. Successive applications of such morphing can be applied to build a path of images that connects two images $\textbf{x}_1$ and $\textbf{x}_2$. This ability to gradually morph/transform one sample into another wherein all the intermediate samples along the path are all ``meaningful'' potentially has multiple use-cases. For example, the length of this path can be use as a meaningful measure of image distance for image retrieval. In addition, morphing can also be used for data augmentation or adversarial attacks. We leave our future works as the study of applications of \codename~in security and privacy aspects of machine learning, in particular membership inference and evasion attacks~\cite{shokri2017membership, evade_ccs}.

%% file: paper.bbl
\begin{thebibliography}{19}
\providecommand{\natexlab}[1]{#1}
\providecommand{\url}[1]{\texttt{#1}}
\expandafter\ifx\csname urlstyle\endcsname\relax
  \providecommand{\doi}[1]{doi: #1}\else
  \providecommand{\doi}{doi: \begingroup \urlstyle{rm}\Url}\fi

\bibitem[Chang et~al.(2008)Chang, Ratinov, Roth, and
  Srikumar]{importance_semantic}
Chang, Ming-Wei, Ratinov, Lev-Arie, Roth, Dan, and Srikumar, Vivek.
\newblock Importance of semantic representation: Dataless classification.
\newblock In \emph{AAAI}, volume~2, pp.\  830--835, 2008.

\bibitem[Dang et~al.(2017)Dang, Huang, and Chang]{evade_ccs}
Dang, Hung, Huang, Yue, and Chang, Ee-Chien.
\newblock Evading classifiers by morphing in the dark.
\newblock In \emph{Proceedings of the 2017 ACM SIGSAC Conference on Computer
  and Communications Security}, pp.\  119--133. ACM, 2017.

\bibitem[Denton et~al.(2015)Denton, Chintala, Fergus, et~al.]{denton2015deep}
Denton, Emily~L, Chintala, Soumith, Fergus, Rob, et~al.
\newblock Deep generative image models using a￼ laplacian pyramid of
  adversarial networks.
\newblock In \emph{Advances in neural information processing systems}, pp.\
  1486--1494, 2015.

\bibitem[Donahue et~al.(2016)Donahue, Kr{\"a}henb{\"u}hl, and Darrell]{bigan}
Donahue, Jeff, Kr{\"a}henb{\"u}hl, Philipp, and Darrell, Trevor.
\newblock Adversarial feature learning.
\newblock \emph{arXiv preprint arXiv:1605.09782}, 2016.

\bibitem[Dumoulin et~al.(2016)Dumoulin, Belghazi, Poole, Lamb, Arjovsky,
  Mastropietro, and Courville]{ali}
Dumoulin, Vincent, Belghazi, Ishmael, Poole, Ben, Lamb, Alex, Arjovsky, Martin,
  Mastropietro, Olivier, and Courville, Aaron.
\newblock Adversarially learned inference.
\newblock \emph{arXiv preprint arXiv:1606.00704}, 2016.

\bibitem[Goodfellow et~al.(2014)Goodfellow, Pouget-Abadie, Mirza, Xu,
  Warde-Farley, Ozair, Courville, and Bengio]{gan}
Goodfellow, Ian, Pouget-Abadie, Jean, Mirza, Mehdi, Xu, Bing, Warde-Farley,
  David, Ozair, Sherjil, Courville, Aaron, and Bengio, Yoshua.
\newblock Generative adversarial nets.
\newblock In \emph{Advances in neural information processing systems}, pp.\
  2672--2680, 2014.

\bibitem[Griffiths et~al.(2007)Griffiths, Steyvers, and
  Tenenbaum]{topics_semantic}
Griffiths, Thomas~L, Steyvers, Mark, and Tenenbaum, Joshua~B.
\newblock Topics in semantic representation.
\newblock \emph{Psychological review}, 114\penalty0 (2):\penalty0 211, 2007.

\bibitem[Huang et~al.(2017)Huang, Li, Poursaeed, Hopcroft, and
  Belongie]{stacked_gan}
Huang, Xun, Li, Yixuan, Poursaeed, Omid, Hopcroft, John, and Belongie, Serge.
\newblock Stacked generative adversarial networks.
\newblock In \emph{IEEE Conference on Computer Vision and Pattern Recognition
  (CVPR)}, volume~2, pp.\ ~4, 2017.

\bibitem[Ioffe \& Szegedy(2015)Ioffe and Szegedy]{batch_norm}
Ioffe, Sergey and Szegedy, Christian.
\newblock Batch normalization: Accelerating deep network training by reducing
  internal covariate shift.
\newblock In \emph{International conference on machine learning}, pp.\
  448--456, 2015.

\bibitem[Jaiswal et~al.(2017)Jaiswal, AbdAlmageed, Wu, and Natarajan]{bcgan}
Jaiswal, Ayush, AbdAlmageed, Wael, Wu, Yue, and Natarajan, Premkumar.
\newblock Bidirectional conditional generative adversarial networks.
\newblock \emph{arXiv preprint arXiv:1711.07461}, 2017.

\bibitem[Kingma \& Welling(2013)Kingma and Welling]{vae}
Kingma, Diederik~P and Welling, Max.
\newblock Auto-encoding variational bayes.
\newblock \emph{arXiv preprint arXiv:1312.6114}, 2013.

\bibitem[Li \& Luo(2017)Li and Luo]{li_gan}
Li, Zhigang and Luo, Yupin.
\newblock Generate identity-preserving faces by generative adversarial
  networks.
\newblock \emph{arXiv preprint arXiv:1706.03227}, 2017.

\bibitem[Lipton \& Tripathi(2017)Lipton and Tripathi]{lipton2017precise}
Lipton, Zachary~C and Tripathi, Subarna.
\newblock Precise recovery of latent vectors from generative adversarial
  networks.
\newblock \emph{arXiv preprint arXiv:1702.04782}, 2017.

\bibitem[Makhzani et~al.(2015)Makhzani, Shlens, Jaitly, Goodfellow, and
  Frey]{aae}
Makhzani, Alireza, Shlens, Jonathon, Jaitly, Navdeep, Goodfellow, Ian, and
  Frey, Brendan.
\newblock Adversarial autoencoders.
\newblock \emph{arXiv preprint arXiv:1511.05644}, 2015.

\bibitem[Mirza \& Osindero(2014)Mirza and Osindero]{cgan}
Mirza, Mehdi and Osindero, Simon.
\newblock Conditional generative adversarial nets.
\newblock \emph{arXiv preprint arXiv:1411.1784}, 2014.

\bibitem[Perarnau et~al.(2016)Perarnau, van~de Weijer, Raducanu, and
  {\'A}lvarez]{icgan}
Perarnau, Guim, van~de Weijer, Joost, Raducanu, Bogdan, and {\'A}lvarez,
  Jose~M.
\newblock Invertible conditional gans for image editing.
\newblock \emph{arXiv preprint arXiv:1611.06355}, 2016.

\bibitem[Radford et~al.(2015)Radford, Metz, and Chintala]{dcgan}
Radford, Alec, Metz, Luke, and Chintala, Soumith.
\newblock Unsupervised representation learning with deep convolutional
  generative adversarial networks.
\newblock \emph{arXiv preprint arXiv:1511.06434}, 2015.

\bibitem[Salakhutdinov \& Larochelle(2010)Salakhutdinov and
  Larochelle]{generative_DBM}
Salakhutdinov, Ruslan and Larochelle, Hugo.
\newblock Efficient learning of deep boltzmann machines.
\newblock In \emph{Proceedings of the Thirteenth International Conference on
  Artificial Intelligence and Statistics}, pp.\  693--700, 2010.

\bibitem[Shokri et~al.(2017)Shokri, Stronati, Song, and
  Shmatikov]{shokri2017membership}
Shokri, Reza, Stronati, Marco, Song, Congzheng, and Shmatikov, Vitaly.
\newblock Membership inference attacks against machine learning models.
\newblock In \emph{Security and Privacy (SP), 2017 IEEE Symposium on}, pp.\
  3--18. IEEE, 2017.

\end{thebibliography}
